\documentclass[letterpaper, 10 pt, conference]{ieeeconf}  

\IEEEoverridecommandlockouts                              

\usepackage{booktabs}
\usepackage{kotex} 
\usepackage{amsmath,amssymb}
\usepackage{graphicx}
\usepackage[caption=false,font=footnotesize]{subfig} 
\usepackage{amsmath,amssymb}
\usepackage[caption=false,font=footnotesize]{subfig} 
\usepackage{pifont}                                  
\newcommand{\cmark}{\ding{51}}                       
\usepackage{algorithm}
\usepackage{algorithmic}
\usepackage{makecell}
\title{\LARGE \bf
OASIS-DC: Generalizable Depth Completion via Output‑level Alignment of Sparse‑Integrated Monocular Pseudo Depth
}

\author{Jaehyeon Cho$^{1}$ and Jhonghyun An$^{1}$%
\thanks{$^{1}$Vehicle Intelligence Perception Lab (VIPLAB), Gachon University, Seongnam-si, Republic of Korea.}%
\thanks{E-mail: jjh000503@gachon.ac.kr, jhonghyun@gachon.ac.kr}%
}

\begin{document}

\maketitle
\thispagestyle{empty}
\pagestyle{empty}

\begin{abstract}
Recent monocular \emph{foundation} models excel at zero-shot depth estimation, yet their outputs are inherently \emph{relative} rather than metric, limiting direct use in robotics and autonomous driving. We leverage the fact that relative depth preserves global layout and boundaries: by calibrating it with sparse range measurements, we transform it into a pseudo \emph{metric} depth prior. Building on this prior, we design a refinement network that follows the prior where reliable and deviates where necessary, enabling accurate metric predictions from very few labeled samples. The resulting system is particularly effective when curated validation data are unavailable, sustaining stable scale and sharp edges across few-shot regimes. These findings suggest that coupling foundation priors with sparse anchors is a practical route to robust, deployment-ready depth completion under real-world label scarcity.

\end{abstract}

\section{INTRODUCTION}

\noindent\textbf{Depth completion}---the task of inferring dense, metric depth from sparse range measurements, optionally guided by RGB---is a key enabler for robust perception and planning in robotics and autonomous driving. Despite rapid advances on the KITTI benchmark~\cite{Uhrig2017KITTI,geiger2012kitti} and strong RGB--LiDAR fusion methods~\cite{Tang2021TIP,Qiu2019DeepLidar,PENet2021ICRA}, prevailing pipelines tacitly assume access to large labeled sets and substantial validation curation. In deployed settings, however, operating conditions (city, route, sensor rig) change faster than labels can be collected; models must generalize from very few samples and with minimal validation. At the same time, recent \emph{foundation-guided} approaches tend to \emph{couple} depth completion to high-dimensional features of a monocular foundation depth estimator (MDE)---e.g., prompting or feature extraction inside the MDE backbone~\cite{Park2024DepthPrompting,ParkJeon2024UniDC}. Such feature-level coupling increases runtime memory/latency and complicates deployment. 

We address these challenges with a \emph{prior-guided} framework that operates at the \emph{output level}: rather than consuming foundation features, we use only the dense depth \emph{output} of a generalizing MDE (e.g., ZoeDepth~\cite{Bha2023ZoeDepth} or Depth Anything~\cite{Yang2024DepthAnything}) and \emph{align} it with sparse range anchors (e.g., LiDAR) to form a calibrated pseudo-depth prior. Concretely, we correct the MDE prediction with available sparse points to obtain a metrically accurate, edge-preserving pseudo map via a Poisson formulation with hard constraints, and feed this prior (plus an anchor mask) to a lightweight refinement network that learns only a residual. The model is designed to follow the prior where consistent, yet deviate via residual corrections in regions of mismatch (thin structures, textureless regions, mixed-depth boundaries). This output-level pairing leverages the appearance generalization of foundation MDEs while sparse anchors fix absolute scale and local geometry, \emph{shrinking the hypothesis space} and preserving few-shot stability \emph{without} the overhead of feature-level coupling. Empirically, this design yields a compact learnable core that is favorable for deployment. 

We evaluate primarily on KITTI Depth Completion~\cite{Uhrig2017KITTI}---our target deployment domain for autonomous driving---and additionally verify indoor generalization on NYUv2~\cite{Silberman2012NYU}. To assess both comparability and deployment realism, we adopt two complementary settings: a \emph{standard few-shot} evaluation and a \emph{strict, deployment-oriented} regime in which training and evaluation share the same few-shot budget. Full dataset splits and protocols are detailed in §IV. This evaluation probes whether reliable generalization is attainable in the absence of a large, curated validation set. 

\noindent\textbf{Contributions.} (1) We introduce a \emph{nonlearned}, MDE-guided pseudo-depth construction that aligns a foundation MDE to sparse anchors and reconstructs a metrically accurate, edge-preserving prior via a Poisson formulation with hard constraints---requiring no trainable parameters. (2) Building on this pseudo prior, we propose a lightweight prior-guided depth completion network that ingests the pseudo map (and anchor mask) and learns only a residual refinement under few-shot supervision, yielding sharp boundaries and robust performance under severe data scarcity---\emph{all while avoiding feature-level coupling to the foundation MDE} for practical deployability.

\section{Related Work}

\begin{figure*}[t]
  \centering
  \includegraphics[width=\textwidth]{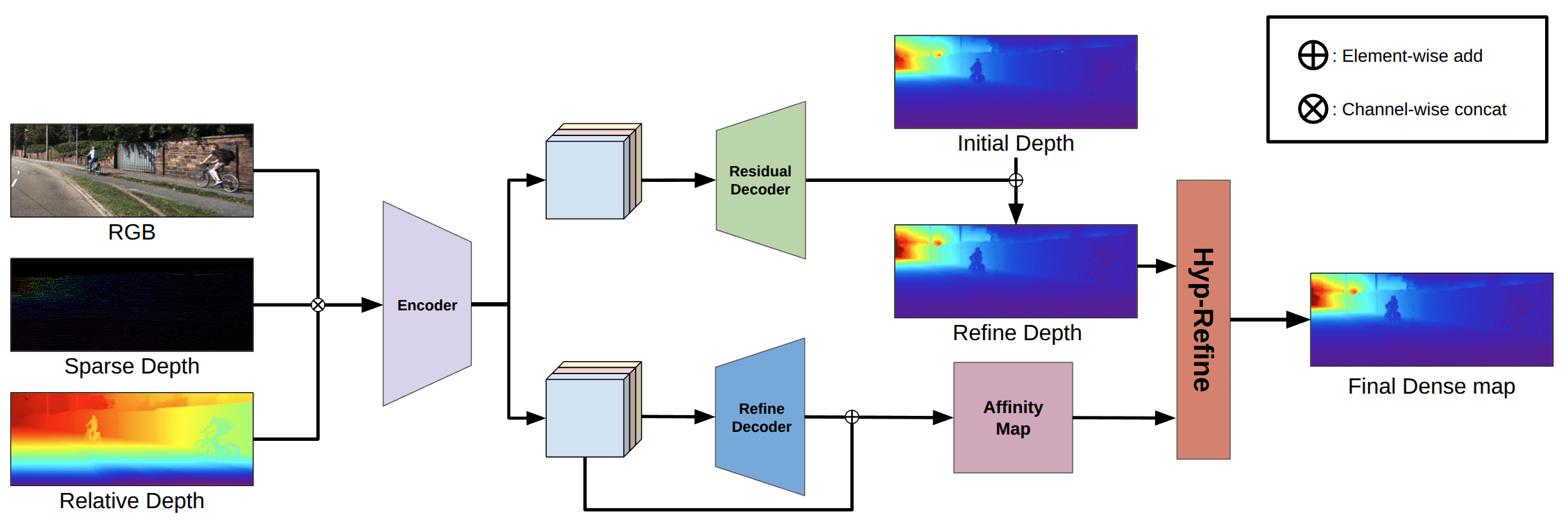}
  \caption{\textbf{Pipeline overview.} RGB, sparse depth, and relative (monocular) depth are concatenated and encoded. A residual decoder predicts an \emph{initial} depth anchored to sparse points, while a refinement branch estimates an \emph{affinity map} for edge/structure–aware propagation. The two streams are fused and passed through a hyperbolic refinement module (Hyp-Refine) to yield the final dense map. Symbols: $\oplus$ element-wise add, $\otimes$ channel-wise concat.}
  \label{fig:mcpropnet_overview}
\end{figure*}


\subsection{Traditional depth completion.}
Image-guided depth completion from sparse range measurements has progressed via RGB--LiDAR fusion and learned affinity/propagation within encoder--decoder frameworks~\cite{Park2020NLSPN,Cheng2018CSPN,Lin2022DySPN}. 
While these approaches attain strong accuracy on curated benchmarks such as KITTI Depth Completion~\cite{Uhrig2017KITTI, geiger2012kitti}, they typically presume abundant pixel-wise supervision and a fixed sparsity pattern, which limits robustness when sensor characteristics, collection routes, or environments shift. 
In practice, dependence on large training/validation splits and tuning to a specific LiDAR density/pattern often impairs cross-sensor and cross-domain generalization.

\subsection{Few-shot depth completion and deployment constraints.}
In deployed robotics/AutonomousDriving(AD) settings, operating conditions (city, route, rig) evolve faster than labels can be curated; models must generalize from very few labeled samples and with minimal validation. 
Leaderboard-style protocols typically rely on full training splits and extensive validation curation on KITTI~\cite{Uhrig2017KITTI,geiger2012kitti}. 
Complementarily, we emphasize deployment-oriented few-shot evaluation that probes generalization under strict data scarcity---e.g., restricting both training and evaluation to the same few-shot subsets and reporting performance on a compact, fixed validation set (KITTI provides an official 1,000-frame validation split)~\cite{Uhrig2017KITTI}. 
Such protocols directly test whether stable metric scale and sharp boundaries are attainable without large validation curation.

\subsection{Foundation-guided completion.} Recent work leverages foundation MDEs to reduce data requirements and improve cross-sensor generalization. DepthPrompting\cite{Park2024DepthPrompting} introduces a depth-prompt module that encodes sparse depth and fuses it with image features to construct pixel-wise affinity, embedding the prompt into a pre-trained MDE to mitigate sensor-range/pattern biases and steer predictions toward absolute-scale depth with lightweight bias tuning. UniDC \cite{ParkJeon2024UniDC} defines universal depth completion and presents a simple baseline that (i) extracts depth-aware features from a foundation MDE, (ii) aligns arbitrary sparse measurements via a pixel-wise affinity built on high-resolution foundation features, and (iii) embeds learned features in a hyperbolic space to capture hierarchical 3D structure, thereby improving zero-/few-shot adaptation across sensors \cite{Park2024DepthPrompting,ParkJeon2024UniDC} . Both lines capitalize on broad appearance priors from foundation MDEs such as ZoeDepth and Depth Anything \cite{Bha2023ZoeDepth}, \cite{Yang2024DepthAnything} and illustrate the value of coupling them with sparse anchors. While effective, such \emph{feature-level} coupling to the foundation MDE backbone~\cite{Park2024DepthPrompting,ParkJeon2024UniDC} often increases runtime memory/latency and complicates deployment. In contrast, we operate at the \emph{output level}: we use only the dense depth \emph{output} of a frozen MDE to build a calibrated pseudo-depth prior anchored by sparse points.

\section{Method}

\subsection{Pseudo-Depth Map Construction}
\label{sec:pseudo_depth}

To stabilize training and model selection under limited-data conditions, we construct a \emph{pseudo-depth} map \(P\) by fusing a dense monocular prior \(E \in \mathbb{R}^{H\times W}\) with sparse LiDAR depth \(D \in \mathbb{R}^{H\times W}\).
\begin{figure}[!t]
  \centering
  \includegraphics[width=\columnwidth]{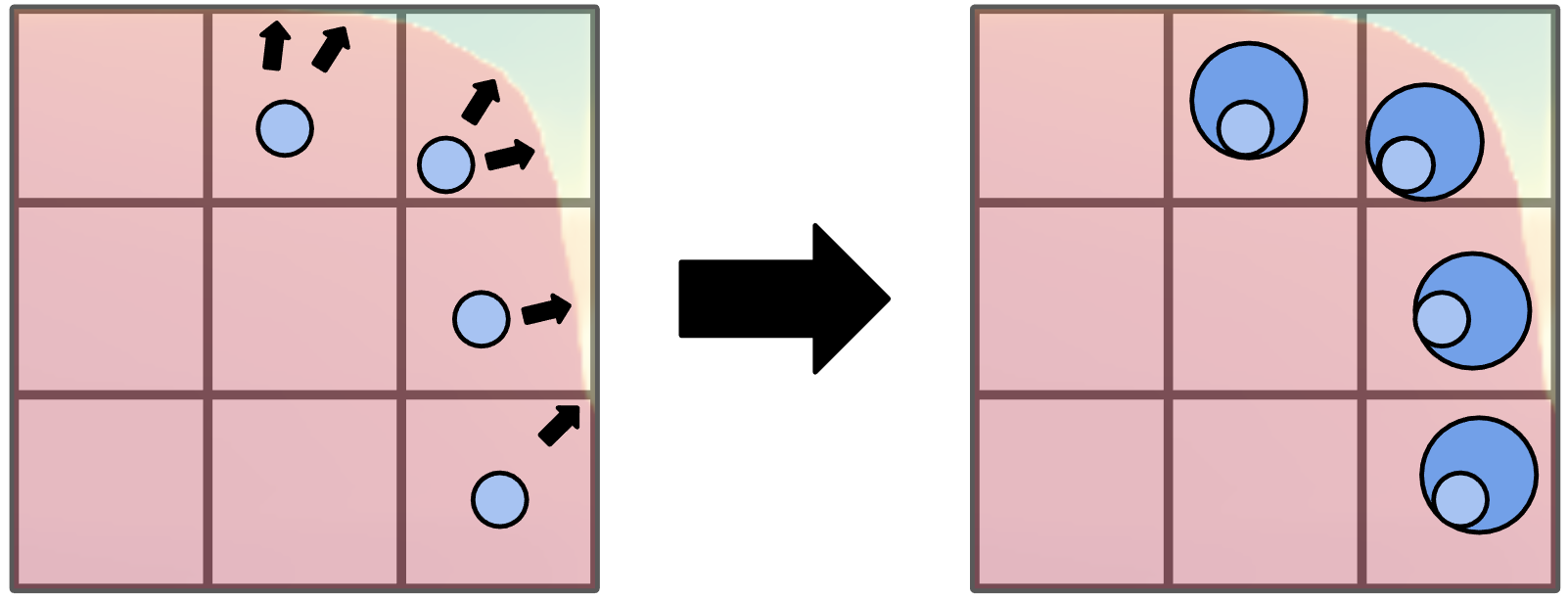}
  \caption{\textbf{Gradient-guided densification.}
    The aligned prior induces a smooth gradient field (background), while sparse metric anchors (blue) and the image boundary are kept fixed.
    Depth values are propagated along the prior’s gradient directions to fill unknown cells, preserving structure and avoiding cross-edge bleeding; anchors are not moved—only the missing pixels are completed.}
  \label{fig:pseudo_depth_densification}
\end{figure}

\paragraph{Sets and boundary data.}
Let $\Omega$ denote the image domain with boundary $\partial\Omega$. We first define the set of known pixels $\mathcal{K}$ and its complement $\mathcal{U}$:
\begin{equation}
\begin{aligned}
\mathcal{K} &= \{(i,j)\in \Omega \mid D_{ij} > 0\} \cup \partial\Omega,\\
\mathcal{U} &= \Omega \setminus \mathcal{K}.
\end{aligned}
\label{eq:sets}
\end{equation}
Based on these sets, we define the Dirichlet field $v_{\text{known}}(i,j)$ used for hard constraints:
\begin{equation}
v_{\text{known}}(i,j) =
\begin{cases}
D_{ij}, & (i,j) \in \{D>0\},\\
E_{ij}, & (i,j) \in \partial\Omega.
\end{cases}
\label{eq:v_known}
\end{equation}

\paragraph{Gradient-domain guidance.}
Following gradient-domain fusion~\cite{Perez2003Poisson}, we align the unknown solution $x$ to the guidance gradients $\nabla E$ while exactly honoring measurements and boundary values:
\begin{equation}
\min_{x}\int_{\Omega}\|\nabla x - \nabla E\|^{2}\,dp 
\quad \text{s.t. } x|_{\mathcal{K}} = v_{\text{known}}.
\label{eq:energy}
\end{equation}
The Euler--Lagrange condition yields a Poisson equation on the unknown set $\mathcal{U}$:
\begin{equation}
\begin{aligned}
A_{\mathcal{U}\mathcal{U}}\, x_{\mathcal{U}} &= \big(L E - L v_{\text{known}}\big)_{\mathcal{U}}.
\end{aligned}
\end{equation}

\begin{equation}
\begin{aligned}
P &= x = x_{\mathcal{U}} + v_{\text{known}}.
\end{aligned}
\label{eq:lin_sys}
\end{equation}
Equations~\eqref{eq:energy}--\eqref{eq:lin_sys} define a strictly convex problem with a unique solution.

\begin{equation}
P(p)=x(p)=
\begin{cases}
v_{\text{known}}(p), & p\in\mathcal{K},\\
x_{\mathcal{U}}(p), & p\in\mathcal{U}.
\end{cases}
\label{eq:reconstruct}
\end{equation}

\paragraph{Discretization and solver.}

On the pixel grid we use the (negative) discrete Laplacian matrix $L$. To preserve discrete consistency, we take the right-hand side as $L E$ (the discrete Laplacian of $E$) and restrict the linear system to the unknown index set $\mathcal{U}$, yielding \eqref{eq:lin_sys}; reconstruction follows \eqref{eq:reconstruct}. Here $A_{\mathcal{U}\mathcal{U}}$ is $L$ restricted to $\mathcal{U}$. The system is symmetric positive definite (SPD), enabling efficient solution via conjugate gradients (CG)~\cite{Hestenes1952CG}. We implement $A_{\mathcal{U}\mathcal{U}}$ in a matrix-free manner by masking a single $3\times 3$ convolution that realizes $L$; each CG iteration requires $O(HW)$ work and only image-sized auxiliary memory.

\subsection{Residual Correction for Monocular Prior}
\label{sec:res}
\paragraph{Motivation.}
Under limited-data training with scarce validation, purely data-driven predictors tend to exhibit high variance and unstable convergence, especially around thin structures and depth discontinuities. We therefore adopt a \emph{prior-preserving residual update}: a reliable but imperfect monocular prior provides the global metric scaffold, while a low-capacity decoder corrects only localized errors. This reduces the effective hypothesis space and improves sample efficiency without sacrificing scene-scale consistency.

\paragraph{Input representation and encoder.}
We construct the input tensor $X$ by spatially aligning and concatenating the following channels:
\begin{equation}
X = \big[I,\; P,\; E,\; M_L\big].
\label{eq:xfeat}
\end{equation}
where $I$ is the RGB image; $P$ is the pseudo-depth prior obtained by Poisson fusion of a monocular estimate with LiDAR anchors; $E$ is the dense monocular prior; and $M_L$ is the binary LiDAR observation mask. Prior to concatenation, we normalize both the pseudo depth and the monocular prior to the unit range using the dataset maximum depth:
\[
P \leftarrow \mathrm{clip}\!\left(\tfrac{P}{d_{\max}},0,1\right),\quad
E \leftarrow \mathrm{clip}\!\left(\tfrac{E}{d_{\max}},0,1\right).
\]
With slight abuse of notation, we continue to denote the normalized maps by \(P\) and \(E\).
All losses are computed in this normalized space.
At inference, predictions are rescaled back to meters via \(D = d_{\max}\,\hat{D}\) and, if needed, clamped to \([0, d_{\max}]\). A vision encoder $\Phi$ produces multi-modal features $F=\Phi(X)$. We intentionally constrain the encoder capacity to mitigate overfitting and stabilize optimization in the few-shot regime.

\paragraph{Residual decoder and prior-preserving update.}
A shallow residual decoder predicts a per-pixel correction $R=f_{\text{dec}}(F)$. The corrected initialization is obtained by adding the residual to the prior:
\begin{equation}
D^{0}(p) = P(p) + R(p).
\label{eq:init}
\end{equation}
(Depths are clamped to $[0, d_{\max}]$ in implementation.) By design, $P$ retains global layout and absolute scale, while $R$ is encouraged to address only local biases near edges, thin structures, and texture-poor regions. The dense monocular prior $E$ is injected via vision encoder and skip connections so that the decoder can allocate residual capacity preferentially at discontinuities, preventing over-smoothing.

\newcommand{\valref}{Sec.~\ref{sec:impl}}
\begin{table*}[t]
\centering
\caption{KITTI Depth Completion Benchmark. Best in \textbf{bold}, second-best is \underline{underlined}. 
\emph{Protocol:} 1/10/100-shot are sampled from the \textbf{training} split only; evaluation is on the \textbf{official 1,000-frame validation} split (no test-server submissions unless stated).}
\label{tab:kitti_depth_completion}

\setlength{\tabcolsep}{4pt}
\renewcommand{\arraystretch}{1.15}
\footnotesize
\begin{tabular*}{\textwidth}{@{\extracolsep{\fill}} l *{8}{c} @{}}
\toprule
 & \multicolumn{2}{c}{1-shot} & \multicolumn{2}{c}{10-shot} & \multicolumn{2}{c}{100-shot} & \multicolumn{2}{c}{1-Sequence Training} \\
\cmidrule(lr){2-3}\cmidrule(lr){4-5}\cmidrule(lr){6-7}\cmidrule(lr){8-9}
Method & RMSE (m) & MAE (m) & RMSE (m) & MAE (m) & RMSE (m) & MAE (m) & RMSE (m) & MAE (m) \\
\midrule
CSPN~\cite{Cheng2018CSPN}             & 9.2748 & 3.5921 & 2.0222 & 0.7825 & 1.4510 & 0.5184 & 2.6289 & 0.8355 \\
S2D~\cite{MaKaraman2018ICRA}          & 8.8479 & 5.6022 & 5.0500 & 3.1469 & 4.2799 & 2.6633 & 4.7950 & 2.5610 \\
NLSPN~\cite{Park2020NLSPN}            & 7.2899 & 4.7422 & 4.0070 & 2.2588 & 2.4979 & 1.1710 & 4.0290 & 1.7881 \\
DySPN~\cite{Lin2022DySPN}             & \underline{2.6350} & \underline{0.8870} & 2.2701 & 0.9150 & 1.8777 & 0.6188 & 2.8530 & 0.7980 \\
CompletionFormer~\cite{CompletionFormer2023} & 4.7212 & 2.3789 & 3.1601 & 1.4740 & 2.6122 & 1.3299 & 4.5588 & 1.9603 \\
BPNet~\cite{Tang2024BPNet}            & 5.4000 & 1.0740 & \underline{1.8799} & \underline{0.5559} & \underline{1.3001} & \underline{0.3910} & \underline{2.1322} & \underline{0.6420} \\
DepthPrompting~\cite{Park2024DepthPrompting} & 2.9840 & 1.1430 & 2.3988 & 1.1290 & 1.8249 & 0.6240 & 2.9468 & 0.9869 \\
Ours                                   & \textbf{1.4190} & \textbf{0.5073} & \textbf{1.2830} & \textbf{0.4001} & \textbf{1.2455} & \textbf{0.3548} & \textbf{1.5782} & \textbf{0.5540} \\
\bottomrule
\end{tabular*}
\end{table*}

\begin{figure*}[t]
    \centering
    \includegraphics[width=\textwidth]{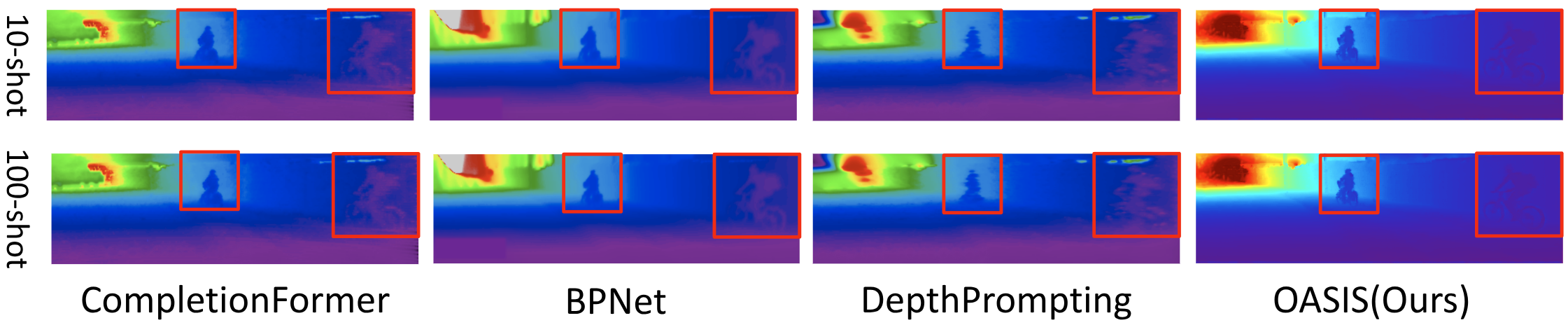}
    \caption{\textbf{KITTI-DC qualitative comparison under few-shot supervision.} 
    Top: 10-shot; bottom: 100-shot. 
    \textbf{Columns:} RGB, CompletionFormer, BPNet, DepthPrompting, and OASIS-DC (Ours).
    Few-shot settings sample only from the training split; visual examples are evaluated against the official 1,000-frame validation set. 
    Red boxes highlight challenging regions (thin structures, far-field). Our results preserve road–wall boundaries and fine details with reduced cross-edge bleeding.}
    \label{fig:mcpropnet_overview1}
\end{figure*}

\subsection{Affinity-Based Refinement}
\paragraph{Scope and inputs.}
Given the initialization $D^{0}\in\mathbb{R}^{H\times W}$ from Sec.~\ref{sec:res}, our goal is to refine it using geometry-aware pixel affinities and a per-pixel sensor anchoring rule that respects LiDAR observations while remaining robust to misprojections. Let $F$ denote encoder features at the image resolution, the raw sensor (LiDAR) depth $D_s$, and the corresponding observation mask $M_L \in \{0,1\}^{H\times W}$. We employ a set of kernel sizes $\mathcal{K}=\{3,5,7\}$ and perform $T$ propagation iterations. A single curvature parameter $\kappa>0$ (shared across $k\in\mathcal{K}$) defines the Poincar\'e ball used for hyperbolic distance evaluations.

\paragraph{Notation.}
For a pixel $p$, $\mathcal{N}_k(p)$ is the $k\times k$ neighborhood; $q\in\mathcal{N}_k(p)$ indexes its neighbors. Kernel gates $\sigma_k(p)\in[0,1]$ satisfy $\sum_{k\in\mathcal{K}}\sigma_k(p)=1$. All row-stochastic affinity weights $A_k(p,q)$ satisfy $\sum_{q\in\mathcal{N}_k(p)} A_k(p,q)=1$.

\paragraph{Hyperbolic pairwise affinity.}
We embed per-pixel features into the Poincar\'e ball~\cite{ganea2018hyperbolic,nickel2017poincare} via the exponential map at the origin:
\begin{equation}
h_p = \exp^{\kappa}_{0}\!\left(W_f F(p)\right).
\label{eq:hyper_embed}
\end{equation}
where $W_f$ is a $1{\times}1$ projection. Let $d_{\kappa}(\cdot,\cdot)$ denote the hyperbolic distance. For each kernel $k\in\mathcal{K}$ and neighbor $q\in \mathcal{N}_k(p)$ we define the unnormalized weights
\begin{equation}
\begin{aligned}
\tilde{A}_k(p,q) &= \exp\!\left(-\frac{d_{\kappa}(h_p,h_q)}{\tau_k}\right)\,\mathbf{1}\{q\in\mathcal{N}_k(p)\},\\
A_k(p,q) &= \frac{\tilde{A}_k(p,q)}{\sum_{q'\in\mathcal{N}_k(p)} \tilde{A}_k(p,q')}.
\end{aligned}
\label{eq:weights}
\end{equation}
with temperature $\tau_k>0$ (per kernel). Pixel-wise kernel gates are obtained by a lightweight head:
\begin{equation}
\sigma_k(p) = \frac{\exp\!\big(g_k(F(p))\big)}{\sum_{k'\in\mathcal{K}} \exp\!\big(g_{k'}(F(p))\big)}, \quad k\in\mathcal{K}.
\label{eq:kernel_gate}
\end{equation}
Intuitively, $3{\times}3$ emphasizes edges and thin structures, $5{\times}5$ aggregates mid-range context, and $7{\times}7$ stabilizes long-range consistency, while the gates adapt these roles per pixel.

\paragraph{Center-tethered multi-kernel propagation.}
To prevent drift and excessive smoothing, each local update is re-centered on $D^{0}$. At iteration $t$, we form a center-tethered patch by replacing the center value with $D^{0}$:
\begin{equation}
\tilde{D}^{(t)}(q) =
\begin{cases}
D^{0}(p), & q=p,\\
D^{(t)}(q), & q\neq p.
\end{cases}
\label{eq:ct_patch}
\end{equation}
Per-kernel propagation and gated mixing are then
\begin{equation}
D^{(t+1)}_{k}(p) = \sum_{q\in\mathcal{N}_k(p)} A_k(p,q)\,\tilde{D}^{(t)}(q),
\label{eq:prop}
\end{equation}
\begin{equation}
D^{(t+1)}_{\text{mix}}(p) = \sum_{k\in\mathcal{K}} \sigma_k(p)\, D^{(t+1)}_{k}(p).
\label{eq:mix}
\end{equation}

\paragraph{Learnable sensor anchoring.}
We enforce per-pixel consistency with sensor observations only where available, while allowing soft corrections near occlusions and misprojections. An anchor map $\alpha(p)\in[0,1]$ is predicted from features:
\begin{equation}
\alpha(p) = \sigma\!\big(W_{\alpha} F(p)\big).
\label{eq:alpha}
\end{equation}
with $\sigma(\cdot)$ the logistic function. At observed pixels ($M_L(p)=1$), the mixed depth in~\eqref{eq:mix} is blended with the raw sensor depth $D_s$:
\begin{equation}
\begin{aligned}
D^{(t+1)}(p) &= \big(1 - \alpha(p)\, M_L(p)\big)\, D^{(t+1)}_{\text{mix}}(p)\\
&\quad +\; \alpha(p)\, M_L(p)\, D_s(p).
\end{aligned}
\label{eq:anchoring}
\end{equation}
and $D^{(t+1)}(p) = D^{(t+1)}_{\text{mix}}(p)$ when $M_L(p)=0$. After $T$ iterations, we output $D_{\text{final}} = D^{(T)}$ and clamp depths to $[0, d_{\max}]$ if necessary.

\section{Experiment}

\begin{table*}[t]
\centering
\caption{NYUv2 Depth Completion Benchmark. Best in \textbf{bold}, second-best is \underline{underlined}.}
\label{tab:nyuv2_depth_completion}
\footnotesize
\setlength{\tabcolsep}{4pt}
\renewcommand{\arraystretch}{1.15}
\begin{tabular*}{\textwidth}{@{\extracolsep{\fill}} l cc cc cc cc}
\toprule
& \multicolumn{2}{c}{1-shot} & \multicolumn{2}{c}{10-shot} & \multicolumn{2}{c}{100-shot} & \multicolumn{2}{c}{1-Sequence Training} \\
\cmidrule(lr){2-3}\cmidrule(lr){4-5}\cmidrule(lr){6-7}\cmidrule(lr){8-9}
Method & RMSE \((\mathrm{m})\) & MAE \((\mathrm{m})\) & RMSE \((\mathrm{m})\) & MAE \((\mathrm{m})\) & RMSE \((\mathrm{m})\) & MAE \((\mathrm{m})\) & RMSE \((\mathrm{m})\) & MAE \((\mathrm{m})\) \\
\midrule
CSPN~\cite{Cheng2018CSPN}             & 1.4827 & 1.2058 & 0.3166 & 0.1961 & 0.2854 & 0.1307 & 0.3166 & 0.1961 \\
NLSPN~\cite{Park2020NLSPN}            & 1.9358 & 1.6132 & 1.5995 & 0.8261 & 0.5501 & 0.4150 & 0.8881 & 0.6421 \\
DySPN~\cite{Lin2022DySPN}             & 1.5474 & 1.2851 & 0.4102 & 0.2817 & 0.3079 & 0.1706 & 0.2584 & 0.1320 \\
CompletionFormer~\cite{CompletionFormer2023} & 1.8218 & 1.5539 & 1.1583 & 1.0162 & 0.9914 & 0.8164 & 0.6779 & 0.5356 \\
CostDCNet~\cite{CostDCNet}            & 1.2298 & 0.9754 & 0.2363 & 0.1288 & 0.1770 & 0.0836 & 0.2066 & 0.0954 \\
BPNet~\cite{Tang2024BPNet}            & 0.3573 & 0.2077 & 0.2392 & 0.1120 & 0.1757 & 0.0793 & 0.2220 & 0.1040 \\
DepthPrompting~\cite{Park2024DepthPrompting} & 0.3583 & 0.2067 & 0.2195 & 0.1006 & 0.2101 & 0.1008 & 0.2335 & 0.1191 \\
\textbf{UniDC}~\cite{ParkJeon2024UniDC}   & \textbf{0.2099} & \textbf{0.1075} & \textbf{0.1657} & \textbf{0.0794} & \textbf{0.1473} & \textbf{0.0669} & \textbf{0.1632} & \textbf{0.0745} \\
Ours                                   & \underline{0.2105} & \underline{0.1105} & \underline{0.1670} & \underline{0.0838} & \underline{0.1484} & \underline{0.0706} & \underline{0.1644} & \underline{0.0787} \\
\bottomrule
\end{tabular*}
\end{table*}

\begin{figure*}[t]
  \centering
  \includegraphics[width=\textwidth]{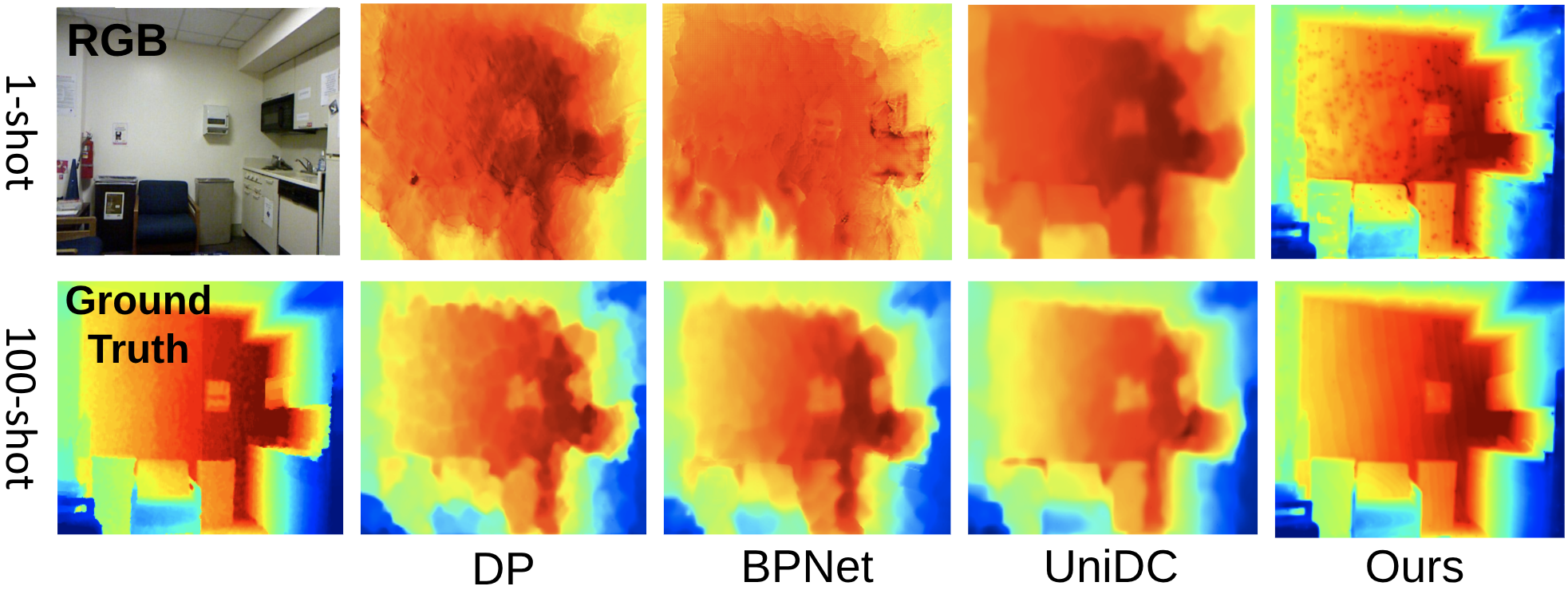} 
  \caption{\textbf{NYUv2 qualitative comparison.} Top row: 1-shot (left: RGB); bottom row: 100-shot (left: Ground Truth). Remaining columns show DP (DepthPrompting), BPNet, UniDC, and Ours. The proposed method produces cleaner planar surfaces and sharper discontinuities (e.g., cabinet edges), while suppressing noise and texture copying across shots.}
  \label{fig:nyuv2_qual}
\end{figure*}

\subsection{Datasets \& Evaluation.}
We evaluate on two complementary benchmarks that span outdoor driving and indoor scenes, and report results under their official protocols.

\paragraph{KITTI Depth Completion (KITTI\mbox{-}DC).}
A large-scale outdoor driving dataset with synchronized RGB images and LiDAR measurements from a Velodyne HDL\textendash64E. 
Following the official split, we use approximately 86K training samples, 7K validation samples, and 1K test samples.
Images are provided at a resolution of 1216$\times$352. 
The raw sparse LiDAR depth covers roughly 6\% of image pixels, while the benchmark’s reference depth is produced via \emph{multi-sweep accumulation and cleanup}, resulting in about 20\% density.
\emph{Evaluation.} Unless otherwise noted, we report quantitative results on the official validation split; submissions to the test server are used only when explicitly stated. Few-shot subsets (1\text{-}/10\text{-}/100\text{-}shot and 1\text{-}Sequence) are drawn from the training split, and validation/test labels are never used for training.

\paragraph{NYUv2.}
A canonical indoor RGB\textendash D dataset captured with a Microsoft Kinect sensor across diverse scenes such as offices, kitchens, and living spaces. 
We adopt the standard labeled subset at 640$\times$480 resolution and follow the conventional official split for training and evaluation. 
For depth completion, sparse inputs are synthesized by sub-sampling dense depth maps to LiDAR-like sparsities; the same sparsity masks are shared across methods to ensure fairness.
\emph{Evaluation.} As in KITTI\mbox{-}DC, unless otherwise noted we report on the official validation split and do not access validation/test annotations during training; few-shot regimes mirror the KITTI settings.

\paragraph{Metrics.}
We report RMSE (meters) and MAE (meters). Lower is better for both. 
All metrics are computed on valid ground-truth pixels using each dataset’s official protocol, with invalid/unknown pixels excluded from the averages.

\subsection{Implementation Details.}
All experiments are executed on a single NVIDIA RTX A5000 (24\,GB) GPU. We adopt a strict few-shot protocol on the official training splits of standard depth-completion benchmarks (e.g., KITTI-DC~\cite{Uhrig2017KITTI,geiger2012kitti} and NYUv2~\cite{Silberman2012NYU}): from each dataset we construct \emph{1/10/100-shot} subsets as well as a \emph{1-sequence} subset consisting of one contiguous sequence. For each shot level and for the sequence setting, sampling is repeated with \(r\) independent random seeds; unless otherwise noted we report the mean (and, when space permits, the standard deviation) over these \(r\) trials. Training uses only the few-shot subsets, and validation/test annotations are never used for training. As the foundation monocular depth estimator, we use \emph{Depth Anything v2}~\cite{yang2024depth}.

Optimization is conducted strictly in few-shot mode, with the number of iterations scaled to subset size: from \(\sim\!100\) iterations for 1-shot up to \(3{,}000\) iterations for larger few-shot subsets (e.g., 100-shot) and the 1-sequence case. Unless otherwise specified, all other hyperparameters are kept fixed across shots to ensure comparability.
\noindent\textbf{Loss functions.}
We supervise with two masked objectives. Let $D_{gt}$ denote ground-truth depth and
\begin{equation}
\begin{aligned}
M &= \mathbb{1}[D_{gt} > 0],\\
n &= \max\!\big(1, \sum M\big).
\end{aligned}
\label{eq:mask}
\end{equation}
The composite L1+L2 loss is
\begin{equation}
\begin{aligned}
e &= (D_{\text{pred}} - D_{gt}) \odot M,\\
\mathcal{L}_{\text{L1+L2}} &= \frac{1}{n}\sum \big(|e| + e^{2}\big).
\end{aligned}
\label{eq:l1l2}
\end{equation}
The scale-invariant log loss operates on positive predictions $D_{\text{rel}}^{+}$ (we use a normalized input in $(0,1]$ with $\varepsilon$-clamping):
\begin{equation}
\begin{aligned}
d &= \big(\log(D_{\text{rel}}^{+}+\varepsilon)
\\[-2pt]&\phantom{=} - \log(D_{gt}+\varepsilon)\big)\odot M,\\
\mathcal{L}_{\text{SI-Log}} &= \frac{1}{n}\sum d^{2}
\;-\; \Big(\tfrac{1}{n}\sum d\Big)^{2}.
\end{aligned}
\label{eq:silog}
\end{equation}
If $D_{gt}$ is unavailable for a sample, the corresponding term is set to zero. Unless stated otherwise, we sum the two losses.

\subsection{Quantitative Result.}
\paragraph{KITTI.}
The improvements in Table~\ref{tab:kitti_depth_completion} stem from how our model decomposes the problem into \emph{metric alignment} and \emph{local residual correction}. 
Sparse LiDAR anchors fix the global, low-frequency structure (absolute scale and large smooth surfaces), which a few labeled samples alone cannot reliably estimate in the 1/10/100-shot regimes. 
On top of this calibrated pseudo-depth, the residual branch concentrates its limited capacity on high-frequency errors around discontinuities. 
This design directly targets the \emph{penetration} failure mode (depth leaking across object boundaries): the residual is learned to deviate from the prior specifically where gradients disagree with the measurements, which sharpens boundaries and suppresses cross-edge bleeding. 
The ablations corroborate this mechanism: using only residuals leaves metric drift unresolved; using only scale calibration reduces bias but cannot repair edge-local artifacts; combining both yields consistent gains across shots. 
That these gains persist from 1-shot to 1-Sequence indicates the model’s inductive bias—not more supervision—is the primary driver: the prior constrains the hypothesis space, while the residual acts as a targeted, data-efficient corrector rather than a full predictor.

\paragraph{NYUv2.}
As shown in Table~\ref{tab:nyuv2_depth_completion}, our method is consistently second-best, narrowly trailing UniDC while outperforming the remaining baselines. This gap is explained by two factors: (i) \emph{weaker anchoring} on NYUv2—Kinect GT and synthesized sparse masks provide less reliable constraints near discontinuities than real LiDAR on KITTI—limiting how strongly metric alignment can regularize the prior; and (ii) an \emph{intentionally conservative residual capacity} for few-shot stability that can underfit fine indoor details relative to UniDC’s larger heads. Even so, the qualitative comparisons in Fig.~\ref{fig:nyuv2_qual} show sharper edges and cleaner planes than other networks, consistent with the aggregated metrics in Table~\ref{tab:nyuv2_depth_completion}.

\begin{figure}[h]
  \centering
  \includegraphics[width=\columnwidth]{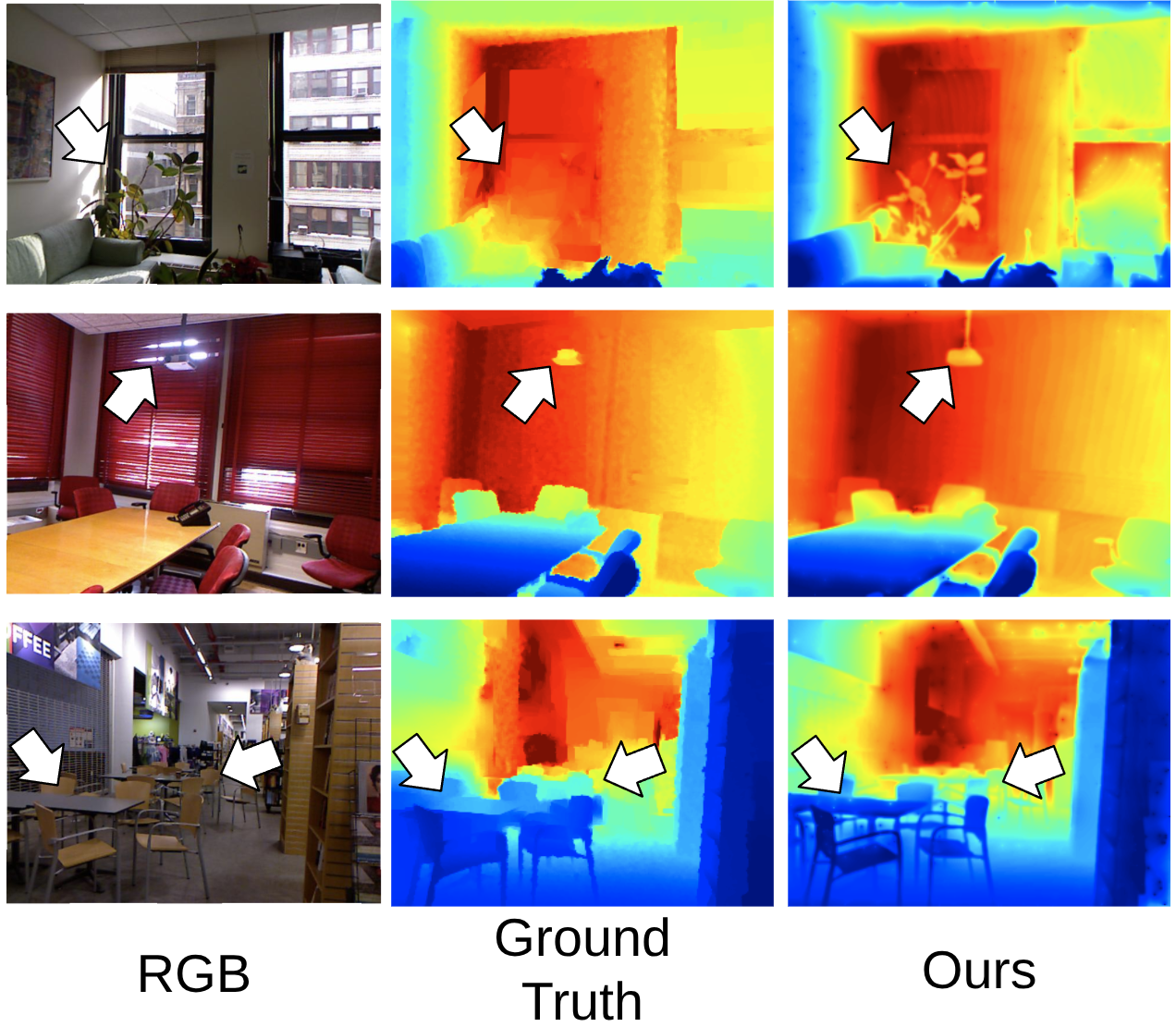} 
  \caption{\textbf{Imperfect NYUv2 ground truth.} Arrows mark GT artifacts (holes/smoothing); metrics use valid GT masks.}
  \label{fig:nyuv2_imperfect_gt}
\end{figure}
\begin{table}[tbp]
\centering
\caption{KITTI Depth Completion Benchmark (ablation).
Residual/Scale flags denote module usage: (\checkmark,–) residual head only; (–,\checkmark) pseudo-prior (scale) only; (\checkmark,\checkmark) both; (–,–) neither.}
\label{tab:kitti_dc_ablation}
\small
\setlength{\tabcolsep}{4pt}
\renewcommand{\arraystretch}{1.2}
\resizebox{\columnwidth}{!}{%
\begin{tabular}{l c cc cc cc}
\toprule
\multicolumn{8}{c}{{KITTI Depth Completion Benchmark}} \\
\midrule
\multicolumn{2}{c}{} & \multicolumn{2}{c}{{1-shot}} & \multicolumn{2}{c}{{10-shot}} & \multicolumn{2}{c}{{100-shot}} \\
\cmidrule(lr){3-4} \cmidrule(lr){5-6} \cmidrule(lr){7-8}
{Residual} & {Scale} & {RMSE} & {MAE} & {RMSE} & {MAE} & {RMSE} & {MAE} \\
\midrule
 &        & 2.8584 & 2.1554 & 2.2840 & 1.5562 & 2.2371 & 1.5109 \\
 \cmark      &        & 4.4062 & 2.7533 & 3.5208 & 1.9879 & 3.4485 & 1.9300 \\
       & \cmark & 2.8939 & 2.1404 & 2.3140 & 1.5469 & 2.2666 & 1.5009 \\
\cmark & \cmark & \textbf{1.4190} & \textbf{0.5073} & \textbf{1.2830} & \textbf{0.4001} & \textbf{1.2455} & \textbf{0.3548} \\
\bottomrule
\end{tabular}%
}
\end{table}
\subsection{Ablation Study.}

\begin{figure}[t]
  \centering
  \includegraphics[width=\linewidth]{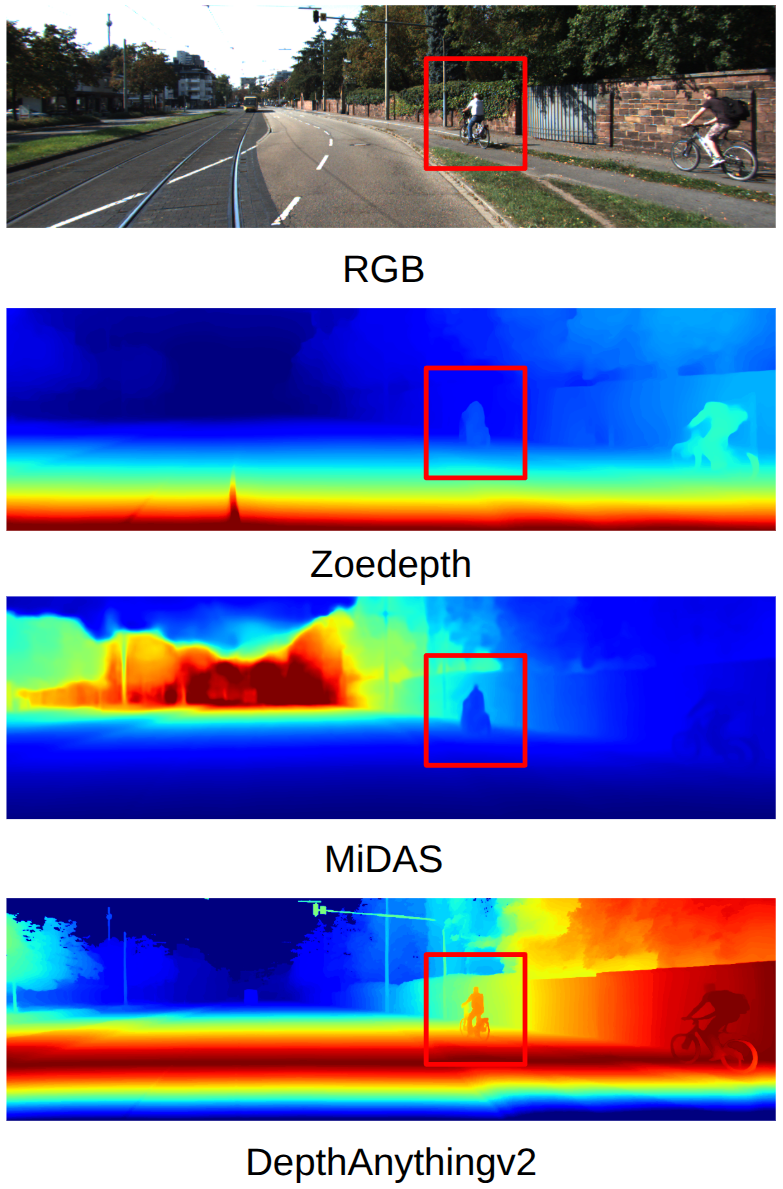}
  \caption{\textbf{DepthAnythingv2 excels at boundary preservation.}
  Qualitative comparison on a KITTI scene. Within the red boxes, DepthAnythingv2 preserves fine object boundaries
  and thin structures more reliably than ZoeDepth and MiDaS, yielding sharper silhouettes and fewer color bleeding artifacts.}
  \label{fig:da2_boundary}
\end{figure}

\begin{table}[tbp]
\centering
\caption{Ablation on pseudo-depth priors on KITTI and NYUv2. Pseudo-depth maps are obtained by test-time densification from frozen foundation MDEs without any learned refinement.}\label{tab:kitti_nyuv2_ours}
\small
\setlength{\tabcolsep}{6pt}
\renewcommand{\arraystretch}{1.2}
\resizebox{\columnwidth}{!}{%
\begin{tabular}{l cc cc}
\toprule
& \multicolumn{2}{c}{{KITTI}} & \multicolumn{2}{c}{{NYUv2}} \\
\cmidrule(lr){2-3}\cmidrule(lr){4-5}
{Method} & {RMSE} & {MAE} & {RMSE} & {MAE} \\
\midrule
Ours\_DepthAnythingv2    & 1.7481 & 0.4525 & \textbf{0.2115} & \textbf{0.1135} \\
Ours\_MiDAS              & 1.7117 & 0.4360 & 0.2741 & 0.1417 \\
Ours\_Zoe                & \textbf{1.6264} & \textbf{0.4135} & 0.3032 & 0.1223 \\
\bottomrule
\end{tabular}%
}
\end{table}

\paragraph{Imperfect Ground Truth}
On NYUv2, our scale–calibrated prior with residual edge refinement yields \emph{second-best} RMSE/MAE across shots (Table~\ref{tab:nyuv2_depth_completion}), narrowly trailing UniDC while surpassing other baselines. Figure~\ref{fig:nyuv2_imperfect_gt} explains the margin: Kinect GT often misses thin structures and over-smooths discontinuities; our residual head sharpens boundaries and recovers small objects, which can increase pixel error in boundary bands despite visibly cleaner geometry. In ablations (prior-only / residual-only / full), only the full model jointly stabilizes global scale and suppresses cross-edge penetration, producing the most reliable trade-off between quantitative scores and edge fidelity.

\begin{table}[t]
\centering
\caption{Computational cost of models.}
\label{tab:comp_cost}
\scriptsize 
\setlength{\tabcolsep}{3pt} 
\renewcommand{\arraystretch}{1.12}
\begin{tabular*}{\columnwidth}{@{\extracolsep{\fill}} l c c c c}
\toprule
Model & \makecell{Total\\Param.} & \makecell{Learnable\\Param.} & \makecell{Infer.\\Time (s)} & \makecell{GPU\\Memory\\(MiB)} \\
\midrule
BPNet \cite{Tang2024BPNet}               & 89.874M  & 89.874M  & 0.072 & 4792 \\
LRRU \cite{Wang2023LRRU}                & \textbf{20.843M}  & 20.843M  & 0.038 & 3650 \\
CompletionFormer \cite{CompletionFormer2023}    & 83.574M  & 83.574M  & 0.060 & 4206 \\
\midrule
Ours               & 94.550M & \textbf{0.219M}   & \textbf{0.013} & \textbf{1904} \\
\bottomrule
\end{tabular*}
\end{table}

\paragraph{Ablation Study of Network Design.}
Table~\ref{tab:kitti_dc_ablation} and Table~\ref{tab:comp_cost} jointly indicate that our residual head operates in an extreme low-capacity regime (\(0.219\)M learnable parameters, \(0.013\)s/inference; Table~\ref{tab:comp_cost}). In this setting, \emph{anchors} (scale calibration from sparse depth) are critical: without anchors, the residual-only variant underfits and amplifies monocular prior biases, yielding substantially worse errors than even the \emph{neither} configuration across 1/10/100-shot. By contrast, when both modules are disabled, the pseudo-depth prior alone imposes a strong inductive bias that “floors” performance (Table~\ref{tab:kitti_dc_ablation}, first row), preventing catastrophic drift despite the tiny trainable head. The best results arise when anchors and residual are combined (last row): anchors stabilize global metric scale and low-frequency structure, while the residual allocates its limited capacity to high-frequency edge corrections (penetration suppression, boundary sharpening). The pattern is consistent across shot regimes, underscoring that with a compact learnable core (Table~\ref{tab:comp_cost}) the anchor signal is not merely helpful but \emph{necessary} for reliable depth completion; the residual then provides targeted refinement rather than bearing the burden of learning scale from few examples.

\paragraph{Ablation on pseudo-depth priors.}
Table~\ref{tab:kitti_nyuv2_ours} evaluates our \emph{training-free} pseudo-depth generation pipeline by swapping only the frozen foundation MDE prior (Depth Anything v2~\cite{yang2024depth}, MiDaS~\cite{Ranftl2022MiDaS}, ZoeDepth~\cite{Bha2023ZoeDepth}) and applying the same test-time densification, with \emph{no learned refinement}. Despite requiring zero training, the method delivers competitive absolute errors across both benchmarks, indicating that the pseudo-depth pipeline \emph{alone} is sufficiently strong for practical deployment when labeled data or training budgets are constrained.

On KITTI~\cite{Uhrig2017KITTI,geiger2012kitti}, the Zoe prior attains the best numbers (RMSE/MAE $=$ 1.6264/0.4135), while the Depth Anything v2 prior reports 1.7481/0.4525---about $\sim$7.5\% higher RMSE and $\sim$9.4\% higher MAE relative to Zoe. We caution that KITTI provides \emph{sparse} LiDAR supervision, and point-sampled metrics tend to favor smoother predictions or those coinciding with LiDAR sampling; in our qualitative inspections, Depth Anything v2 preserves object boundaries and thin structures more crisply, which is not fully reflected by the sparse metrics. Consistently, on the \emph{dense} indoor NYUv2 benchmark~\cite{Silberman2012NYU}, Depth Anything v2 achieves the best errors (0.2115/0.1135), improving over MiDaS by $\approx$22.8\%/19.9\% and over ZoeDepth by $\approx$30.2\%/7.2\% in RMSE/MAE, respectively. These results underscore that our training-free pseudo-depth maps are already robust enough for industrial use, and that dense ground-truth evaluation better captures the edge fidelity exhibited by the Depth Anything v2 prior.

\section{CONCLUSIONS}

We presented a prior-guided few-shot depth completion framework that fuses a foundation MDE with sparse anchors to form a calibrated pseudo-depth prior and trains a refinement network to respect the prior while correcting local mismatches. This complementary design reduces the effective hypothesis space, stabilizes learning from very few samples, and preserves both metric scale and edge detail. Our evaluation on KITTI adheres to two complementary protocols: a standard few-shot setting for comparability and a strict, deployment-oriented regime that uses five random seeds over 1-/10-/100-shot and 1-sequence subsets (70/30 train/val splits except 1-shot), with final reporting on the official 1{,}000-image validation split that is never used for training or hyperparameter tuning~\cite{Uhrig2017KITTI,geiger2012kitti}. Ablations disentangling the roles of the pseudo prior and sparse anchors (no-prior / no-sparse / full-prior) indicate that removing the prior markedly harms generalization while removing anchors increases scale drift. Future work includes stronger uncertainty-aware calibration of the monocular prior, dynamic-scene handling, and extensions to varying LiDAR sparsities and cross-dataset generalization.

\addtolength{\textheight}{-12cm}   









\end{document}